\documentclass[10pt,twocolumn,letterpaper]{article}

\usepackage{float}
\usepackage[pagenumbers]{cvpr} 

\usepackage{booktabs}
\usepackage{lipsum}
\usepackage{graphicx}
\usepackage{soul}
\usepackage[skip=1.2pt]{caption}
\usepackage{colortbl}
\usepackage[dvipsnames]{xcolor}

\newcommand\kr[1]{{\color{green}[KR]: #1}}

\definecolor{cvprblue}{rgb}{0.21,0.49,0.74}
\usepackage[pagebackref,breaklinks,colorlinks,citecolor=cvprblue]{hyperref}

\makeatletter
\renewcommand{\paragraph}{%
  \@startsection{paragraph}{4}%
  {\z@}{0.1ex \@plus 0.2ex \@minus .2ex}{-0.5em}%
  {\normalfont\normalsize\bfseries}%
}

\def\paperID{6739} 
\def\methodName{\texttt{ShotAdapter}}
\def\confName{CVPR}
\def\confYear{2025}

\title{{ShotAdapter: Text-to-Multi-Shot Video Generation with Diffusion Models}}

\author{Ozgur Kara$^{1,2}$\qquad Krishna Kumar Singh$^2$\qquad
Feng Liu$^2$\qquad 
Duygu Ceylan$^2$\qquad \\
James M. Rehg$^1$\qquad Tobias Hinz$^2$
\\
{$^1$UIUC}\qquad
{$^2$Adobe}
\\
Project Webpage: \href{https://shotadapter.github.io/}{https://shotadapter.github.io/}}

\begin{document}

\twocolumn[{%

\renewcommand\twocolumn[1][]{#1}%
\maketitle
\vspace{-3em}
\begin{center}
    \centering
    \captionsetup{type=figure}
    \includegraphics[width=0.9\textwidth]{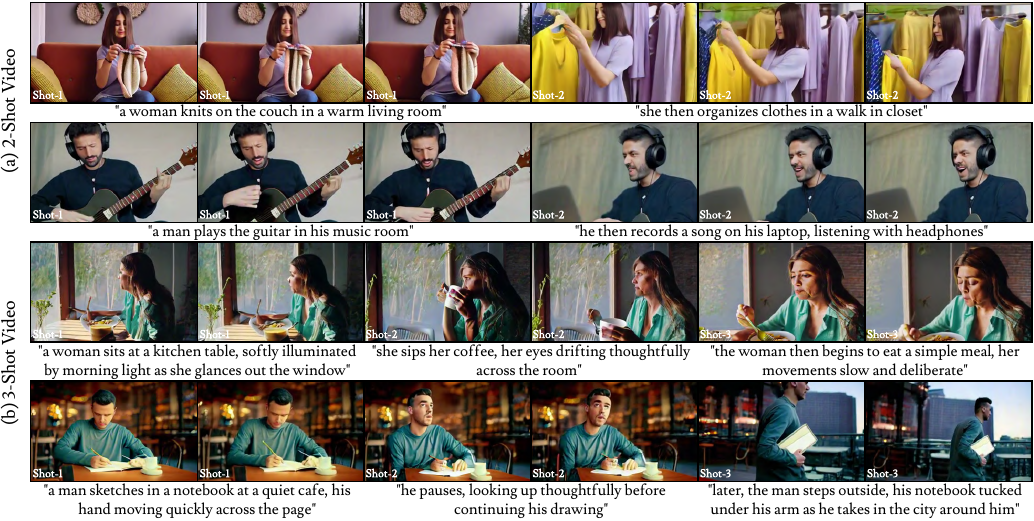}
    \captionof{figure}{\textbf{\methodName}\ is a lightweight framework that enables text-to-multi-shot video generation by fine-tuning a pre-trained text-to-video model. It allows control over number and duration of shots as well as shot content through shot-specific text prompts. The framework maintains character identity while being able to preserve backgrounds (\eg 3\textsuperscript{rd} row) or transition to new ones (\eg 4\textsuperscript{th} row), featuring distinct activities (\eg playing guitar, then using laptop) and perspectives.}
    \label{fig:teaser}
\end{center}%
}]

\maketitle

\begin{abstract}
Current diffusion-based text-to-video methods are limited to producing short video clips of a single shot and lack the capability to generate multi-shot videos with discrete transitions where the same character performs distinct activities across the same or different backgrounds. To address this limitation we propose a framework that includes a dataset collection pipeline and architectural extensions to video diffusion models to enable text-to-multi-shot video generation. Our approach enables generation of multi-shot videos as a single video with full attention across all frames of all shots, ensuring character and background consistency, and allows users to control the number, duration, and content of shots through shot-specific conditioning. This is achieved by incorporating a transition token into the text-to-video model to control at which frames a new shot begins and a local attention masking strategy which controls the transition token's effect and allows shot-specific prompting. To obtain training data we propose a novel data collection pipeline to construct a multi-shot video dataset from existing single-shot video datasets. Extensive experiments demonstrate that fine-tuning a pre-trained text-to-video model for a few thousand iterations is enough for the model to subsequently be able to generate multi-shot videos with shot-specific control, outperforming the baselines. You can find more details in \href{https://shotadapter.github.io/}{our webpage}.
\end{abstract}

\section{Introduction}
\label{sec:intro}

\begin{figure}[t]
\centering
\includegraphics[width=\columnwidth]{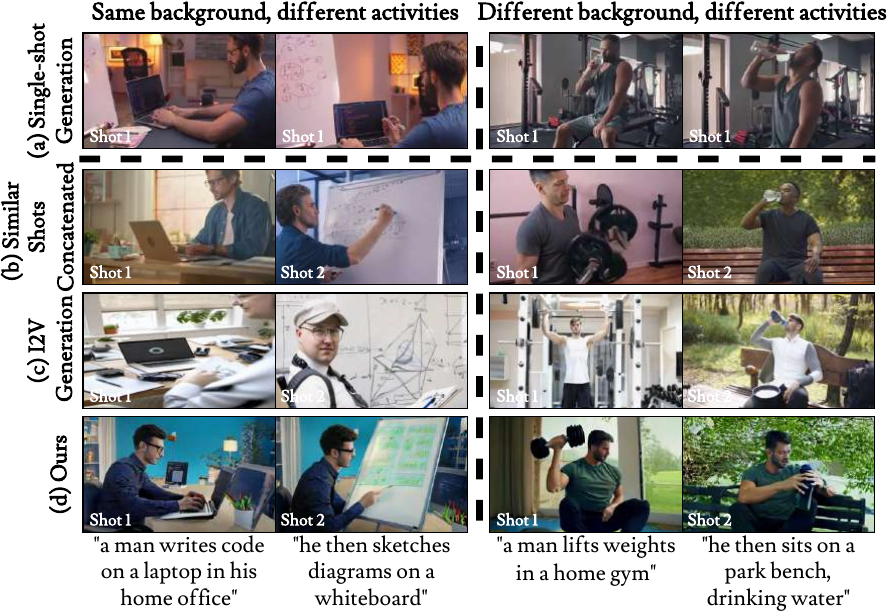}
    \caption{\textit{\textbf{Comparison with baselines.}} Each row displays frames from generated 2-shot videos, guided by shot-specific text prompts. The left column shows results with same background and different activities, while the right column presents results both with different backgrounds and activities.}
    \label{fig:mot}
\end{figure}

While diffusion models~\cite{rombach2022high, ho2020denoising} have shown impressive capabilities in the image domain~\cite{ruiz2023dreambooth, wei2023elite,galimage,ye2023ip, ruiz2024hyperdreambooth,hertz2024style,zheng2023layoutdiffusion,bansal2023universal,podellsdxl, ramesh2022hierarchical}, extending them to video synthesis presents significant challenges due to the dynamic nature of videos. 
One group of researchers has explored various methods to adapt text-to-image (T2I) models for video synthesis~\cite{geyer2024tokenflow, kara2024rave, khachatryan2023text2video, wu2023tune, singer2023makeavideo}. In contrast, another group has focused on text-to-video (T2V) diffusion models~\cite{brooks2024video, polyak2024movie, jin2024pyramidal} which demonstrate superior performance by processing the entire video as a unified input rather than handling each frame independently. Despite their state-of-the-art performance in video generation, all existing models are designed to generate a single continuous video, which imposes inherent limitations when attempting to generate a multi-shot video of the same character engaged in multiple distinct activities, where each shot is separated by discrete cuts in dynamic elements. This challenge becomes particularly significant when each activity requires a unique setting, even within the same background (Fig.~\ref{fig:mot}, 1\textsuperscript{st} column: a man writes code and then sketches diagrams on a whiteboard) or when the background needs to change while maintaining the same identity (Fig.~\ref{fig:mot}, 2\textsuperscript{nd} column: transitioning from a home gym to a park bench). Moreover, existing models are limited to generating videos of very short durations, and therefore lack flexibility.  Because they have been trained exclusively on single-shot videos, they are ill-suited for multi-shot video synthesis. Consequently, these limitations hinder their applicability in real-world contexts, such as film production, where narratives rely on multiple shots, each featuring distinct actions or perspectives and often featuring the same character across diverse scenes and time frames.

The simplest approach to multi-shot synthesis is to generate a single-shot video by combining all shot-specific prompts into a single prompt. However, this method cannot create ``cuts" (transitions between shots) and struggles to provide different backgrounds for the same character. Even within a same background, it cannot generate characters featuring different activities requiring distinct settings, as it is constrained to generating short videos
(Fig.~\ref{fig:mot} (a)). An extension of this approach is to generate each shot many times individually, then concatenate the most similar shots, which requires the generation of numerous videos (Fig.~\ref{fig:mot} (b)). An improvement to this baseline entails first generating consistent keyframes of a character using a reference image, then animating these keyframes with an image-to-video (I2V) model, and finally concatenating the generated clips to produce a multi-shot video. However, this baseline is inevitably limited by the capabilities of off-the-shelf methods, as they struggle with maintaining consistency and quality. Additionally, it remains challenging to depict the character within the same background performing distinct activities (Fig.~\ref{fig:mot} (c)). 

Given the limitations of potential baselines, we introduce \methodName, a simple yet powerful, model-agnostic framework for controllable text-to-multi-shot video (T2MSV) generation. \methodName \space transforms a single-shot T2V generator into a T2MSV generator with minimal fine-tuning, requiring only five thousand iterations (less than 1\% of the total pre-training iterations), on a multi-shot video dataset. This transformation is made possible through a novel attention-layer masking strategy and a special learnable token, called ``transition token", which signals the transition between shots. It enables multi-shot video generation where the camera perspective can shift abruptly between shots (Fig.~\ref{fig:teaser}, 2\textsuperscript{nd} row) within a single background with the character performing different activities, or the background itself can change (Fig.~\ref{fig:teaser}, 4\textsuperscript{th} row), all while preserving the character's identity. Notably, \methodName \space offers users precise control over the content of each generated shot through shot-specific text prompts, along with the flexibility to specify the number and duration of shots. Additionally, the use of a unified input for the T2V model ensures consistency across shots with minimal fine-tuning on an appropriate multi-shot video dataset.

To address the lack of training datasets, we propose two novel pipelines to construct a multi-shot video dataset from a single-shot video dataset through pre- and post-processing steps, which eliminates the need to collect actual multi-shot video.
Recognizing the lack of an evaluation standard for T2MSV generation, we propose an evaluation pipeline and several baseline comparisons to assess synthesis results based on identity and background consistency, text alignment, and shot-duration precision. In summary, our main contributions are as follows:


\begin{itemize}
    \item We propose a model-agnostic and computationally efficient framework that transforms a T2V model into a T2MSV generator with minimal fine-tuning, ensuring the preservation of character identity across all generated shots. 
    
    \item Our approach allows user-defined control over the number of shots, their duration, and the content of each shot through text prompts, leveraging a novel ``transition token" coupled with a localized attention masking strategy.

    \item We propose two novel data collection strategies to construct a multi-shot video dataset derived from single-shot video dataset, along with pre- and post-processing steps.
    
    \item We believe we are the first to frame the T2MSV generation task 
    as a challenge for the research community. Therefore, we introduce an evaluation pipeline to assess performance in this domain and will release a validation dataset to promote standardized evaluation.

\end{itemize}

\section{Related Works}
\label{sec:rel-works}

\paragraph{Text guided video synthesis} 
Text-guided video synthesis is a rapidly evolving field. Early methods adapted pre-trained text-to-image (T2I) diffusion models~\cite{rombach2022high} for video synthesis by modifying architecture, such as adding temporal layers~\cite{wu2023tune}, altering latent inputs~\cite{khachatryan2023text2video}, and introducing temporally correlated noise for frame consistency~\cite{ge2023preserve}. However, T2I models struggle with frame coherence due to their lack of temporal processing. Recent approaches, such as T2V models using unified video inputs and Diffusion Transformer (DiT) frameworks, address this. OpenAI’s SORA~\cite{brooks2024video} and Gen-Tron~\cite{chen2024gentron} enhance DiT with large datasets and advanced text-conditioning. Commercial models like Luma Dream Machine~\cite{DreamMachine_LumaLabs}, MiniMax~\cite{MiniMax}, and Kling AI~\cite{KlingAI} achieve coherence through extensive training. MovieGen~\cite{polyak2024movie} enables instruction-based editing and personalized generation. Open-source contributions have advanced T2V methods, such as W.A.L.T~\cite{gupta2023photorealistic}, which uses a two-stage algorithm for training on image and video datasets, and Latte~\cite{ma2024latte}, which employs Transformer blocks. RIVER~\cite{davtyan2023efficient} and PyramidFlow~\cite{jin2024pyramidal} use autoregressive generation with flow-matching, while OpenSora~\cite{opensora} and OpenSora-Plan~\cite{pku_yuan_lab_and_tuzhan_ai_etc_2024_10948109} aim to replicate SORA more efficiently. In contrast to prior works that focus on single-shot video generation, we introduce a framework for fine-tuning T2V models, transforming them into T2MSV generators.

\paragraph{Image/Video story generation}

Personalized identity-preserving image generation presents challenges in maintaining consistent identity across diverse settings. Techniques like DreamBooth~\cite{ruiz2023dreambooth} and Textual Inversion~\cite{galimage} achieve this through time-consuming test-time fine-tuning. Some line of work ~\cite{xiao2024fastcomposer,wei2023elite,ma2024subject} use adapter models whereas other works~\cite{yan2023facestudio,ye2023ip-adapter,li2024photomaker,wang2024instantid} incorporate ArcFace~\cite{deng2019arcface} or keypoints on face features for conditioning. In image story generation, StoryMaker~\cite{zhou2024storymaker} ensures stylistic consistency, while ConsiStory~\cite{tewel2024training} enhances subject consistency through a shared attention block and feature injection. DreamStory~\cite{he2024dreamstory} leverages Large Language Models, and StoryDiffusion~\cite{zhou2024storydiffusion} introduces an attention mechanism for consistent character representation. In video story generation, DreamBooth~\cite{ruiz2023dreambooth}, adapted in Tune-A-Video~\cite{wu2023tune}, requires fine-tuning for each input video, limiting scalability and restricting outputs to short, single-shot videos based on reference images. Recognizing the lack of a unified solution for multi-shot video generation, one of our baselines combines StoryMaker~\cite{zhou2024storymaker} with an I2V model, though this integration shows limited performance. Our approach achieves superior multi-shot video generation results compared to this baseline.

\begin{figure*}
\centering
\includegraphics[width=0.9\textwidth]{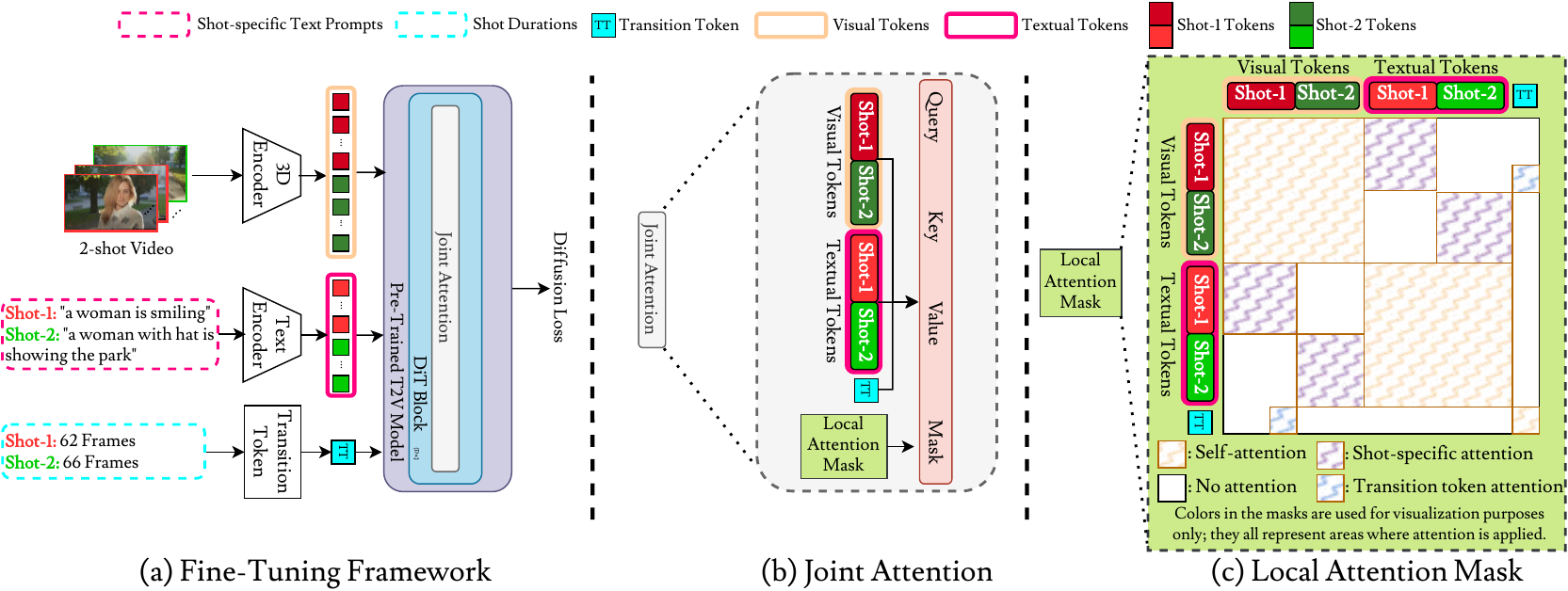}
    \caption{\textit{\textbf{Fine-tuning framework with transition token and local attention masking.}} (a) \methodName \space fine-tunes a pre-trained T2V model by incorporating ``transition tokens" (highlighted in light blue). We use \(n-1\) transition tokens, initialized as learnable parameters, alongside an \(n\)-shot video with shot-specific prompts, which are fed through the pre-trained T2V model. (b) The model processes the concatenated input token sequence, guided by a ``local attention mask" through joint attention layers within DiT blocks. (c) The local attention mask is structured to ensure that transition tokens interact only with the visual frames where transitions occur, while each textual token interacts exclusively with its corresponding visual tokens.}
    \label{fig:model-architecture}
\end{figure*}

\section{Methodology}
\label{sec:method}


\subsection{Preliminaries}

\paragraph{Diffusion formulation} Diffusion models~\cite{sohl2015deep,ho2020denoising,rombach2022high} generate realistic data by learning to reverse a noise-adding process. During training, data is progressively noised over multiple time steps, and the model learns to denoise it step by step, starting from pure noise \(\epsilon \sim \mathcal{N}(0, \mathbf{I})\). A noisy version \(\mathbf{x}_i^{(t)}\) at timestep \(t\) is obtained by adding scaled noise 
\(
\mathbf{x}_i^{(t)} = \sqrt{\alpha_t} \, \mathbf{x}_i + \sqrt{1 - \alpha_t} \, \epsilon,
\)
where \(\alpha_t\) controls the noise level. The training loss
\(
\mathcal{L}_{\text{diffusion}}(\theta) = \mathbb{E}_{\mathbf{x}_i, \epsilon, t}  \left\| \epsilon - \epsilon_{\theta}(\mathbf{x}_i^{(t)}, t) \right\|^2 
\)
minimizes the difference between the true noise \(\epsilon\) and the model’s prediction \(\epsilon_{\theta}(\mathbf{x}_i^{(t)}, t)\) 
This loss trains the model to predict the noise, enabling it to reconstruct the original data through reverse diffusion.

\paragraph{Latent video diffusion models} Given an input video sample \(\mathbf{x}_i \sim p_{data}(\mathbf{x})\) with dimensions \(\mathbf{x}_i \in \mathbb{R}^{F \times C \times H \times W}\), the process begins by encoding the video using a 3D encoder \(\mathcal{E}\), yielding a latent representation \(\mathbf{z}_i \in \mathbb{R}^{F' \times C' \times H' \times W'}\), where \(F' = F / f_t\), \(H' = H / f_s\), and \(W' = W / f_s\), with \(f_s\) and \(f_t\) being the spatial and temporal compression ratios, respectively. The latent video representation is further transformed into a sequence of \(N\) tokens for a DiT-based model, denoted as \(\{\Bar{\mathbf{x}}_i^n\}_{n=1}^{n=N}\), where each token \(\Bar{\mathbf{x}}_i^n \in \mathbb{R}^{D}\) has a hidden dimension \(D\). Patchification is performed along the width, height, and frame depth, patch sizes represented by \(f_{p_w}\), \(f_{p_h}\), and \(f_{p_f}\), respectively.

\subsection{Text-to-Multi-Shot Video Generation}

Inline with the definitions in previous works~\cite{zhang1993automatic,chasanis2008scene,thompson2003film}, a \textbf{shot} is defined as the smallest segment of a video—a single, uninterrupted clip with continuous motion. Each shot is characterized by dynamic elements, including the foreground object, background setting, object actions, and camera movement. 
A \textbf{multi-shot video} is defined as a video composed of multiple individual shots, each being separated by a “cut”~\cite{thompson2003film}, which is an instantaneous change in the dynamic elements. More formally, an N-shot video $V$ can be represented as $V=\{s_1^{K_1}, \dots, s_N^{K_N}\}$, where each shot $s_i^{K_i}$ is a set of ${K_i}$ frames, $s_i^{K_i} = \{f_1, \dots, f_{K_i}\}$, with each frame $f_j\in \mathbb{R}^{C\times H\times W}$ representing an image with channel (C), height (H), and width (W). Our work focuses on multi-shot videos featuring a single foreground object, specifically humans, as they are often the main characters and present challenges in maintaining identity consistency for real-world applications. 
We aim to \emph{preserve the identity of the foreground object across shots}, even as it performs different activities. Furthermore, when the background is designated as a fixed location, we require background consistency throughout the entire video. More formally, our task is to generate an N-shot video given a set of input conditions $\{C_1, C_2, ...\}$ where each $C_i$ denotes shot-specific conditions, \eg $C_i = \{\text{shot caption}, \text{shot duration}, \cdots\}$.

\methodName\ introduces model-agnostic novel extensions to transform pre-trained T2V diffusion models into T2MSV generators with minimal fine-tuning along with a multi-shot video dataset curation method. Our method allows users to control the number and duration of each shot as well as the content through shot-specific text prompts.

\begin{figure*}
\centering
\includegraphics[width=0.95\textwidth]{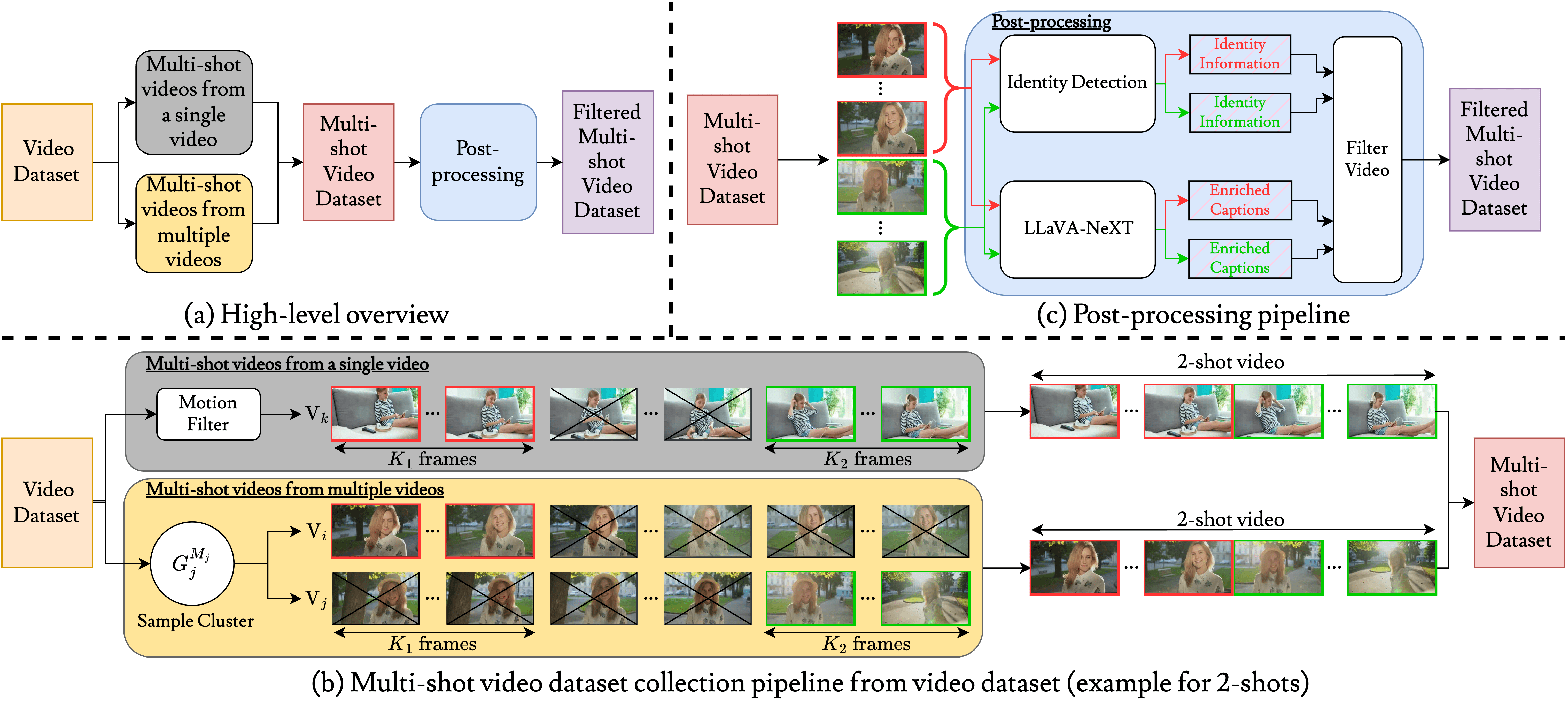}
    \caption{\textit{\textbf{Multi-shot video dataset collection pipeline.}} A high-level overview of this pipeline is presented in (a). Our first method (gray box in (b)) samples videos with large motion, randomly splits them into \( n \)-shots with varied durations, and concatenates them into multi-shot videos. Our second method (yellow box in (b)) randomly samples \( n \) videos from pre-clustered groups containing videos of the same identities and concatenates them to form a multi-shot video. Finally, we post-process (c) the multi-shot videos to ensure identity consistency and obtain shot-specific captions using LLaVA-NeXT~\cite{zhang2024llavanextvideo}. }
    \label{fig:data-collection}
\end{figure*}

\subsection{Model}

\paragraph{Model architecture} We use a diffusion transformer (DiT)~\cite{peebles2023scalable} based T2V framework similar to OpenSora~\cite{opensora} and MovieGen~\cite{polyak2024movie}, shown in Fig.~\ref{fig:model-architecture} (a). Our model integrates a 3D Variational Autoencoder (3D-VAE) for video encoding, along with a variant of joint-attention layers~\cite{esser2024scaling} for conditioning. Specifically, input videos are first subject to temporal and spatial encoding via the 3D encoder before being patchified. The textual condition tokens are then concatenated with these patchified visual tokens and subsequently processed by the DiT. We leverage a pre-trained T2V model that can generate 128 frames with 192$\times$320 resolution, and fine-tune it to enable multi-shot generation. This extension is achieved by introducing a transition token and implementing local attention masking.
Note that our framework also supports cross-attention based conditioning. 


\paragraph{Transition token} 
Inspired by the commonly employed `[EOS]' (End of Sentence) token in natural language processing, which signals the model to recognize the end of a sentence, we propose a novel, learnable embedding referred to as the ``transition token" which enables the model to learn transitions between consecutive shots within a multi-shot video.
Specifically, we initialize a set of learnable parameters at the start of fine-tuning, matching the hidden dimension of the input tokens. We repeat these parameters \(n-1\) times—where \(n\) represents the specified number of shots—and append them to the end of the input visual and textual token sequence (Fig.~\ref{fig:model-architecture} (a)). In the model’s attention layers, we implement a masking strategy that ensures the transition token interacts only with the tokens corresponding to the frames where transitions are intended to occur (Fig.~\ref{fig:model-architecture} (b) and (c)). This approach allows the model to focus on transition frames, enabling it to learn to generate cuts between shots, while also allowing users to specify both the number of shots and their respective durations, by simply replicating the transition token and adjusting the attention mask accordingly.



\paragraph{Local attention masking}


To enable shot-specific control, we introduce a local attention masking technique (Fig.~\ref{fig:model-architecture} (c)). 
Without masking, all tokens interact with each other, diluting the impact of shot-specific information. Our proposed local attention masking strategy overcomes this limitation by restricting attention interactions to specific token groups. Specifically, the attention matrix is masked to enforce the interactions, where the transition token attends only to the tokens at the transition frames, while visual and textual tokens are restricted to self-attention. Additionally, to enhance precise and localized control of textual tokens over their corresponding shot-visual tokens,  each textual token attends solely to its associated visual tokens. Consequently, the fine-tuned model is better equipped to capture fine-grained dependencies between frames, shot-specific text prompts, and shot transitions, resulting in improved control over the T2MSV generation.

\begin{figure*}
\centering
\includegraphics[width=0.95\textwidth]{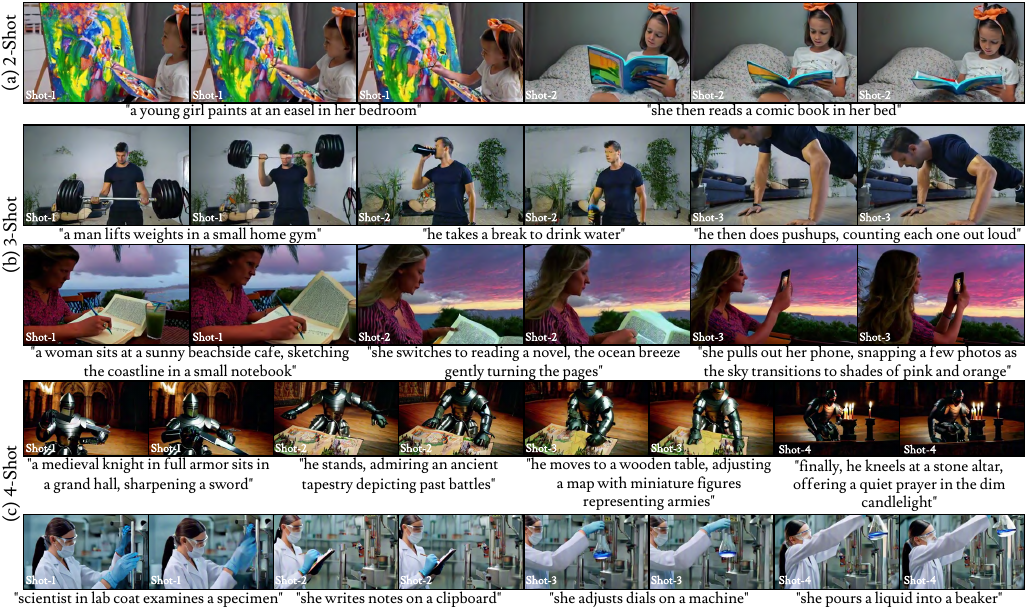}
    \caption{\textit{\textbf{Qualitative results.}} Our approach enables multi-shot video generation depicting different actions and background guided by shot-specific prompts. In the 2\textsuperscript{nd} row, the shots maintain a \textit{consistent background} while capturing different perspectives, whereas the 3\textsuperscript{rd} row depicts the same woman in \textit{related backgrounds} that subtly change in response to the prompt. For complete videos, see Supplementary.}
    \label{fig:qual-results}
\end{figure*}

\subsection{Data}

\begin{figure*}
\centering
\includegraphics[width=\textwidth]{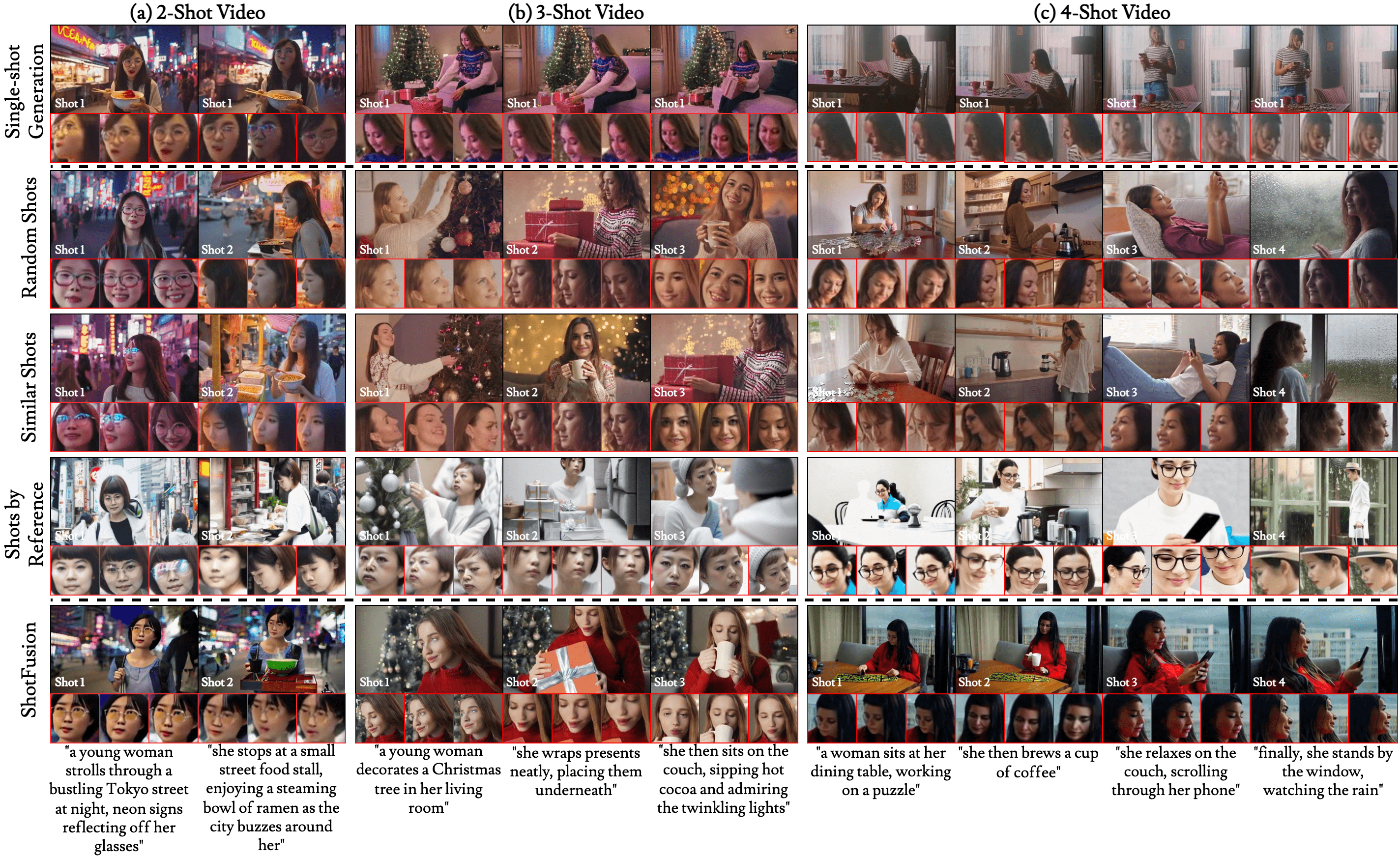}
    \caption{\textit{\textbf{Qualitative comparison.}} We compare our approach to single-shot generation and baseline methods for (a) 2-shot, (b) 3-shot, and (c) 4-shot videos. Our model (last row) enables local control over shot content while Single-Shot Generation fails to feature distinct activities (1\textsuperscript{st} row). Random Shots (2\textsuperscript{nd} row) and Similar Shots (3\textsuperscript{rd} row) struggle to preserve character identity, especially facial features (cropped below frames in red bounding boxes), which our method effectively maintains. Shots by Reference (4\textsuperscript{th} row) improves identity consistency to some extent but falls short in maintaining both identity (\eg shot 4 in (c)) and background coherence (\eg shot 3 in (b)), where our model demonstrates superior performance.
 \textbf{For the full videos, please see  Supplementary.}}
    \label{fig:qual-comp}
\end{figure*}

To perform fine-tuning we require a suitable dataset consisting of multi-shot videos where each shot depicts the same identity both in same and different backgrounds, featuring different activities. We develop two methods to curate multi-shot videos given access to a large-scale video dataset (Fig.~\ref{fig:data-collection} (c)). The first method creates multi-shot videos from long single-shot videos that exhibit large motion while the second method combines multiple independent videos of the same human subjects to produce multi-shot videos. Post-processing is then applied to obtain shot-specific captions and ensure that the foreground identity remains consistent across shots.


\paragraph{Multi-shot videos from single-shot videos}

In this approach we generate a multi-shot video from a single-shot video with large motion by randomly trimming short sub-clips and combining them to create a new n-shot video. (Fig.~\ref{fig:data-collection} (a), gray box). We first obtain optical flow maps using RAFT~\cite{teed2020raft} and compute the average motion in the x and y directions $(t_x, t_y)$ along with scaling ratio $s$ for each video (see Supplementary for details). Videos with motion metrics below specified thresholds are discarded, retaining only those with significant motion in at least one metric. Additionally, videos with fewer than 250 frames are excluded to ensure sufficient variability. Finally, we choose random sub-clips of random duration and concatenate them in a random order to generate a dataset of multi-shot videos.

\paragraph{Multi-shot videos from independent videos}
Despite focusing on large-motion videos, the previous approach encounters limitations in terms of diversity in camera perspective, motion and backgrounds, reducing the variety of generated multi-shot videos. To enhance the dataset variability we perform clustering on the single-video dataset, \ie $\mathcal{D} = \{G_1^{M_1}, \cdots, G_n^{M_n}\}$, where each cluster $G_j^{M_j}$ contains $M_j$ videos featuring the same foreground object. These clusters are constructed by choosing videos uploaded by the same user within the same time-frame (\eg 3 days), and videos that match based on the similarity of their captions. 
we obtain approximately 550K clusters, with an average of ~6 videos per cluster. Subsequently, we generate a fixed number of multi-shot videos for each cluster by randomly trimming clips from separate videos within the same cluster (Fig.~\ref{fig:data-collection} (a), gray-yellow). This methodology significantly enhances the dataset's richness and variability, introducing a wider array of scenes and perspectives.

\paragraph{Post-processing} We apply several post-processing pipelines (Fig.~\ref{fig:data-collection} (c)) to (i) obtain shot-specific prompts and (ii) ensure identity consistency across shots. Since each shot in the multi-shot video dataset requires a specific text description, 
we employ the LLaVA-NeXT~\cite{zhang2024llavanextvideo} model to generate shot-specific captions. Additionally, we apply an additional filtering stage to the multi-shot video dataset to ensure identity consistency across shots. Videos containing more than one character or featuring different identities are filtered out using YOLO~\cite{yolov5} as a person detector in the middle frames of each shot within a multi-shot video. To verify identity consistency, we utilize embeddings from DINOv2 and applied a threshold to ensure the same identity appears across all shots. This post-processing step filters out 38\% of the multi-shot videos, resulting in a dataset composed of multi-shot videos with 2, 3, and 4 shots of varying durations, where a single consistent identity performs similar or different activities across diverse backgrounds and motions.

\section{Experiments}
\label{sec:exp}


Since there is no existing evaluation pipeline for the T2MSV generation task, we design a benchmark to enable standardized evaluations (see Supplementary for implementation details). 
\vspace{-1pt}
\paragraph{Evaluation dataset} 
Our dataset is partitioned according to whether the background remains constant or changes and to the number of shots as specified in the text prompts.  For each multi-shot configuration (2, 3, and 4 shots) we provide 8 prompts per scenario (\ie background remains the same or changes) resulting in a total of 48 prompts for multi-shot settings. These prompts are designed to depict a human subject by including terms such as ``a person" performing different activities in each shot using ChatGPT~\cite{OpenAI_ChatGPT} (see Supplementary). In total, we generate 128 frames for each sample, with randomly selected shot durations.



\paragraph{Baselines} To evaluate our approach in the absence of directly comparable methods we devise three baselines. \textbf{Random Shots (RS)} generates each shot guided by  detailed text descriptions as a single-shot video multiple times (\ie 48)  and concatenates the randomly selected shots. \textbf{Similar Shots (SS)} improves on this by selecting shots based on DINOv2 embedding similarity across foreground objects.  
\textbf{Shots by Reference (SR)} first generates keyframes of consistent characters using StoryMaker~\cite{zhou2024storymaker} and then animates those into individual video shots using our model's I2V capability. Note that for all baselines, we use the original pre-trained model, providing a fair comparison by using the same underlying model architecture.

\paragraph{Quantitative evaluation} 
We adapt commonly used metrics from single-shot video generation~\cite{huang2024vbench}: (a) \textit{Identity Consistency} (IC), which calculates the average DINOv2~\cite{caron2021emerging} embedding similarity between the segmented persons (segmented using YOLO~\cite{yolov5}) at the middle frames of each shot; (b) \textit{Text Alignment} (TA), which assesses the alignment of generated content with text prompts by calculating the similarity between text features and shot features extracted by ViCLIP~\cite{wang2024internvid}, then averaging across shots; and (c) \textit{Background Consistency} (BC), which measures similarity by segmenting the background and computing DINOv2 embedding similarity across the middle frames of each shot. Our approach outperforms all previous baselines in preserving identity for 3- and 4-shots, while achieving competitive results with SR for 2-shots, as measured by the IC metric (Table~\ref{tab:main}). As the number of shots increases, there is a decline in performance due to the propagation and accumulation of errors in generating consistent identities. Additionally, when the background changes (\textit{diff bg}), a modest overall decrease in performance is observed. We report background consistency for samples where the background is intended to remain constant (\textit{same bg}), where our model outperforms all other approaches across every shot by a large margin. In terms of text alignment, RS and SS generally perform better, as expected, since they generate each shot individually, effectively serving as an upper bound for this score. However, our approach achieves competitive results, demonstrating that it preserves identity and background consistency more effectively without significantly compromising text alignment. For additional comparisons to SEINE~\cite{chen2023seine}, MEVG~\cite{oh2024mevg}, FreeNoise~\cite{qiu2024freenoise} and Gen-L-Video~\cite{wang2023genlvideo}, see Supplementary.

\begin{table*}[]
\begin{center}

\caption{\textit{\textbf{Quantitative comparison.}} We evaluate our approach against baseline methods across \textbf{Identity Consistency}, \textbf{Background (BG) Consistency}, \textbf{Text Alignment} and a \textbf{User Study} on 2, 3, and 4-shot videos under conditions with background changes (\textit{diff bg}), without background changes (\textit{same bg}), and with both types (\textit{all}). \methodName\ w/o TT indicates the model fine-tuned without Transition Tokens (TT) and  \methodName\ w/ 2-shots fine-tuned only on a dataset of 2-shot videos. The user study assesses Identity Consistency (Q1), Background Consistency (Q2), and Text Alignment (Q3). $\uparrow$ and $\downarrow$ indicate the direction toward better performance for each metric.}
\label{tab:main}
\resizebox{\textwidth}{!}{%
\begin{tabular}{l|cccccc|ccc|ccc|ccc}
\toprule
 & \multicolumn{6}{c|}{\textbf{Identity Consistency} $\uparrow$} & \multicolumn{3}{c|}{\textbf{BG Consistency} $\uparrow$} & \multicolumn{3}{c|}{\textbf{Text Alignment} $\uparrow$}  & \multicolumn{3}{c}{\textbf{User Study}} \\ \cmidrule{2-16}
\multicolumn{1}{c|}{\textbf{Shot Number}} & \multicolumn{2}{c}{2} & \multicolumn{2}{c}{3} & \multicolumn{2}{c|}{4} & 2 & 3 & 4 & 2 & 3 & 4 & Q1 (IC) & Q2 (BC) & Q3 (TA) \\ 
\multicolumn{1}{c|}{\textbf{Background Change}} & \textit{diff bg} &  \textit{same bg} & \textit{diff bg} & \textit{same bg} & \textit{diff bg} & \textit{same bg} & \multicolumn{3}{c|}{\textit{same bg}}  & \multicolumn{3}{c|}{\textit{all}} & \multicolumn{3}{c}{Ours vs Baseline (Selection ratio in 1-to-1 comparison)} \\ \midrule
Random Shots (RS) & 71.03 & 80.47 & 54.76 & 63.72 & 48.08  & 55.87 &  84.46 & 65.77 & 59.18 &  26.84 & \textbf{26.47} &  \textbf{25.44} & \textbf{77.19}\% / 22.81\% & \textbf{73.27}\% / 26.73\% & \textbf{56.73}\% / 43.27\%
  \\  

Similar Shots (SS) & 73.94  & 82.55 & 55.15 & 66.17   & 49.25 & 58.67 &  88.85    & 67.02 & 60.20 & 26.40  & 26.13 &  25.16 & \textbf{72.92}\% / 27.08\% & \textbf{69.20}\% / 30.80\% & \textbf{53.13}\% / 46.87\% \\ 
Shots by Reference (SR) & \textbf{81.74} & 84.98 & 67.92 & 72.97 & 57.83 & 67.74 & 82.11 & 64.85 & 56.81 & 25.59 & 23.97 & 21.98 & \textbf{73.43}\% / 26.57\% & \textbf{82.28}\% / 17.72\% & \textbf{73.03}\% / 26.97\% \\ \midrule
\methodName \space w/o TT & 77.17 & 84.78 & 68.95 & 70.98  & 58.83 & 70.24 & 87.94 & 72.93 & 70.48 & 26.64 & 23.15 & 22.84 & N/A & N/A & N/A \\ 
\methodName \space w/ 2-shots & 78.05 & 85.46 & 70.12 & 71.53 &  56.99  & 68.37 & 89.08 & 75.53 & 73.19 & 25.97 & 23.59   & 22.97 & N/A & N/A & N/A  \\ \midrule
\methodName & 78.67  & \textbf{86.33} & \textbf{70.30} & \textbf{76.44} & \textbf{61.86}  & \textbf{74.89} &  \textbf{89.48} & \textbf{77.66} & \textbf{76.55} & \textbf{27.12} & 23.65 & 22.17 & N/A & N/A & N/A  \\ \bottomrule

\end{tabular}%
}
\end{center}
\vspace{-8pt}
\end{table*}

\paragraph{User study} To complement the quantitative metrics, we conduct a user study on the Prolific~\cite{Prolific} platform with 75 participants. Each participant views two videos simultaneously, selected from a pool of 10 randomly chosen videos from the generated results, with one video always generated by \methodName. Participants are then asked to choose their preferred video based on identity consistency (IC), background consistency (BC), and text alignment (TA). Our approach achieves superior results in a 1-to-1 comparison with baselines (Table~\ref{tab:main}) in identity and background consistency. In terms of text alignment, it achieves a slight improvement over SS and RS, while outperforming SR by a substantial margin, confirming the trends observed in the quantitative metrics.

\paragraph{Ablation study} We conduct two ablation studies: (i) removing the transition token while retaining the local attention mask (Table~\ref{tab:main} \methodName\ w/o TT), and (ii) fine-tuning the model exclusively on 2-shot videos (Table~\ref{tab:main} \methodName\ w/ 2-shots). Including transition tokens yields a slight improvement in IC, BC, and TA, as it assists the model in generating cuts, thereby enhancing localized control over shot transitions. Although fine-tuning on only 2-shot videos reduces the dataset size, results on 3- and 4-shot videos reveal the model’s generalizability, maintaining better identity and background consistency overall than the baselines, despite a slight performance decrease compared to the final model.

\paragraph{Qualitative evaluation} Fig.~\ref{fig:teaser} and Fig.~\ref{fig:qual-results} show our model’s T2MSV generation capabilities across 2, 3, and 4-shot videos, addressing scenarios that require either background transitions (\eg a transition from a living room to a walk-in closet) or distinct activities within the same setting (\eg a character lifting weights, drinking water, and doing push-ups). Our model effectively generates multi-shot videos of the same characters, with ``cuts” even in diverse settings (full videos are available in the Supplementary). In Fig.~\ref{fig:qual-comp}, we provide a qualitative comparison of our approach with single-shot video generation using extended prompts (row 1) and baseline methods RS, SS and SR for 2, 3, and 4-shot videos. Single-shot generation with localized shot control can result in scenes where actions are intermingled (Fig.~\ref{fig:qual-comp} (a), where the model fails to transition from walking to eating ramen). RS produces the least consistent outcomes, generating characters with random identities (Fig.~\ref{fig:qual-comp}, zoomed-in faces highlighted with red bounding boxes) and incoherent backgrounds due to the random concatenation of shots. SS shows a minor improvement in identity consistency by selecting shots based on foreground similarity, yet still generates visually distinct identities, despite similar clothing (Fig.~\ref{fig:qual-comp} (b)), and struggles to maintain coherent backgrounds. SR, achieves better identity consistency than previous baselines as it generates consistent characters using an off-the-shelf method~\cite{zhou2024storymaker} but suffers from quality degradation as the number of shots increases and lacks temporal coherence between keyframes, resulting in notable inconsistencies, such as complete environment changes (Fig.~\ref{fig:qual-comp} (c)). In contrast, our approach effectively addresses these limitations, generating multi-shot videos with consistent character identity across different background requirements as directed by text prompts.


\begin{table}[t]
\begin{center}
\caption{\textit{\textbf{Transition token generalizability.}} We compute the absolute difference in frames between the generated and ground truth shot duration  as the \textbf{Mean Shot Duration Error (MSDE)}, and the error per-shot is reported with a range of 2 to 8 shots per video.}
\label{tab:ablation}
\resizebox{0.7\columnwidth}{!}{%
\begin{tabular}{lccccccc}
\toprule
 Shots & \textbf{2} & \textbf{3} & \textbf{4} & \textbf{5} & \textbf{6} & \textbf{7} & \textbf{8} \\ \midrule
MSDE & 2.00 & 0.83 & 1.00 & 1.70 & 1.33 & 0.92 & 1.21 \\ \bottomrule
\end{tabular}%
}
\vspace{-7pt}
\end{center}

\end{table}

\paragraph{Transition token generalizability}
To further assess the generalizability of the transition token, we test our model on videos with 2 to 8 shots, using the \textbf{Mean Shot Duration Error} (MSDE) metric, which is calculated by averaging the absolute difference between ground-truth and generated shot durations in terms of frames, where SceneCut~\cite{PySceneDetect} is used to detect the cuts. Quantitative results (Table~\ref{tab:ablation}) show that despite the temporal compression applied during encoding and patchification, the model achieves an average offset of only 1 to 2 frames per shot, even in 8-shot examples. These findings confirm that the transition token serves effectively as an ``End of Shot" marker and can be extended to accommodate multiple shots.

\paragraph{Limitations} While our experiments demonstrate the effectiveness of our approach in T2MSV generation this study is limited to human foreground objects, as experiments with non-human subjects (\eg animals) were not conducted. This limitation is primarily due to dataset filtering choices. Additionally, the maximum duration the model can generate is restricted by the underlying model used for fine-tuning, which is limited to 128 frames in this study. For future work, we aim to extend the duration by employing an autoregressive approach to generate additional shots conditioned on previously generated ones.
It is worth noting that our method experiences a slight quality reduction in the user study compared to baselines, though it remains highly competitive. We hypothesize that this minor drop is primarily due to fine-tuning with a 90\% smaller batch size compared to the baseline as well as our model better adhering to multiple text captions while the baselines often ignore larger parts of the text (see Supplementary for quality analysis).

\section{Conclusion}
In this paper, we present \methodName, a lightweight framework that transforms single-shot T2V models into multi-shot T2MSV generators with minimal fine-tuning. Our approach incorporates a \textit{transition token} and \textit{localized attention masking}, applied to a multi-shot video dataset collected through a novel data collection pipeline. Extensive evaluations demonstrate that our method outperforms baseline models in identity and background consistency without compromising text alignment scores, as further validated by a user study. Additionally, our findings highlight the framework’s generalizability to videos with an increasing number of shots, affirming the effectiveness of the ``transition token" concept.

{
    \small
    \bibliographystyle{ieeenat_fullname}
    \bibliography{main}
}

\clearpage
\setcounter{page}{1}
\renewcommand{\thesection}{S.\arabic{section}}
\setcounter{section}{0}

\maketitlesupplementary

\section{Full Videos}
For the complete videos, please see the HTML file in the supplementary zip.

\section{Quality Analysis}

\begin{figure}[t]
    \centering
    \includegraphics[width=\columnwidth]{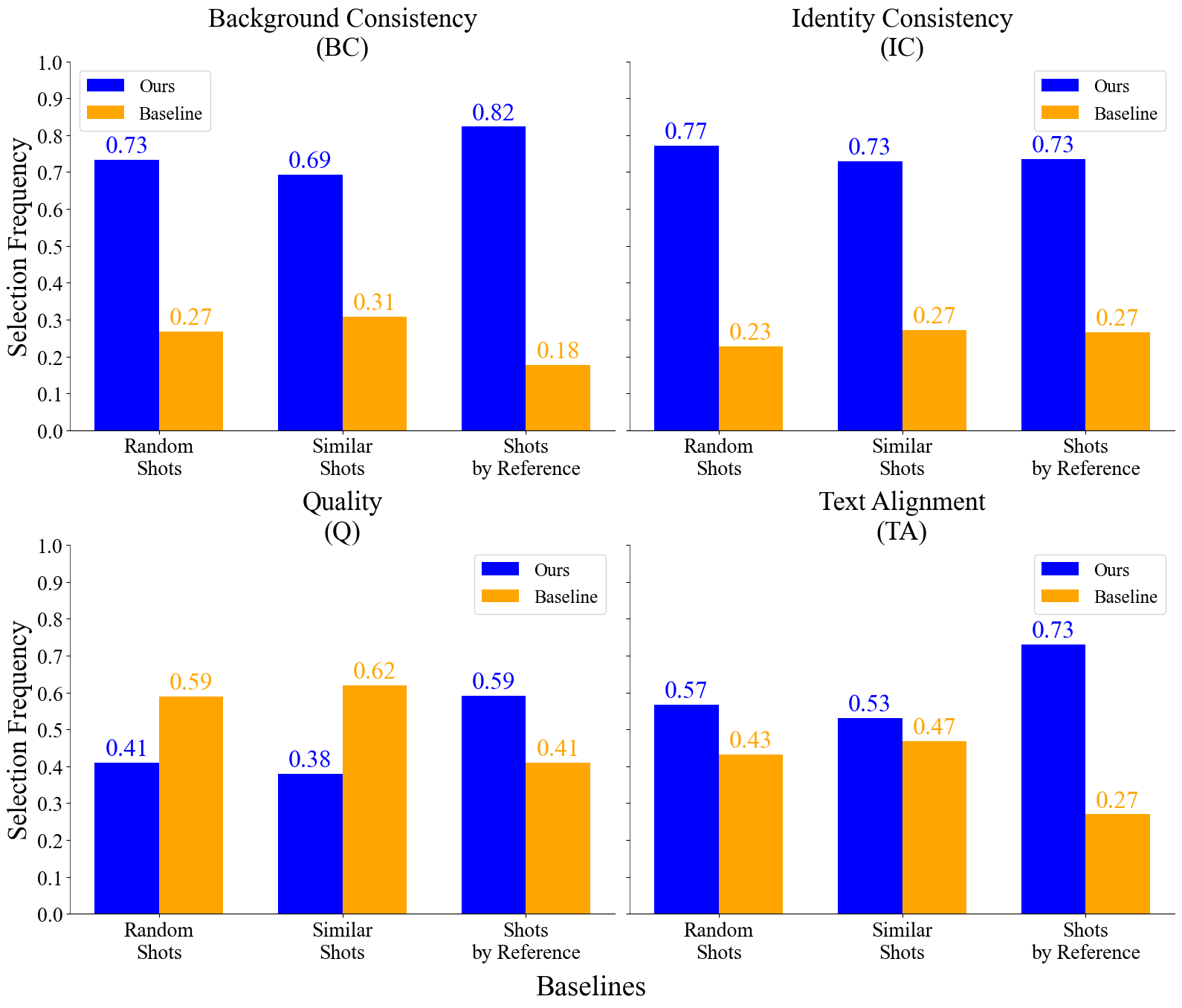}
    \caption{\textbf{\textit{User study results.}}}
    \label{fig:user-study}
\end{figure}

In our user study (Fig.~\ref{fig:user-study}), each participant views two videos simultaneously, selected from a pool of 10 randomly chosen videos from the 
generated results, with one video always generated by ShotAdapter. Participants are then asked to choose their 
preferred video based on Identity Consistency (IC), Background Consistency (BC), Text Alignment (TA), and Quality (Q). 
Our approach achieves superior results in a 1-to-1 comparison with baselines in Identity (IC) 
and Background Consistency (BC). In terms of Text Alignment (TA), it achieves a slight improvement over Similar Shots (SS) and Random Shots (RS), 
while outperforming Shots by Reference (SR) by a substantial margin.
While our model shows slightly lower performance in Quality (Q) compared to RS and SS,
this can be attributed to fine-tuning with a 90\% reduced batch size, emphasizing the lightweight nature of our approach 
as it still brings competitive results despite a training with much reduced batch size. Moreover, our model is shown to be better than SR as using an off-the-shelf method~\cite{zhou2024storymaker} and combining it with I2V model even further propogates the error, resulting in worse quality. Despite this minor trade-off, 
our method excels in all other metrics, demonstrating its robustness and effectiveness for multi-shot video generation.

Although our model demonstrates slightly lower quality compared to the baselines in the user study, 
we hypothesize that this is due to the significantly reduced batch size during fine-tuning. Specifically, we utilize
a batch size that is 90\% smaller than the one used during pre-training. Table~\ref{tab:fvd} illustrates how varying the batch 
size impacts the quality of the generated videos, measured by Frechet Video Distance (FVD)~\cite{unterthiner2019fvd}.
When fine-tuning with a batch size of 32, whether on the original video dataset (Default Dataset) without our data 
collection pipeline or on our processed dataset (ShotAdapter), the FVD scores remain comparable. Notably, our method 
(ShotAdapter) achieves slightly better scores, suggesting that the reduced batch size during fine-tuning, rather 
than our data processing approach, accounts for the measured differences in video quality metric.
Furthermore, our current checkpoint, fine-tuned with a batch size of 128, shows improved quality compared to the 
32 batch size setups. However, it still performs slightly worse than the original checkpoint, which was 
pre-trained with a significantly larger batch size ($>$1000).

\begin{table}[!h]

\begin{center}
\caption{\textit{\textbf{Quality comparison with varying batch size}}}
\label{tab:fvd}
\resizebox{\columnwidth}{!}{%

\begin{tabular}{l|cc|c|c}
\toprule
 & \multicolumn{2}{c|}{\textbf{32 batch size}} & \textbf{\textgreater{}1000 batch size} & \textbf{128 batch size} \\ \cmidrule{2-5}
 & \textit{\begin{tabular}[c]{@{}c@{}}Default\\ Dataset\end{tabular}} & \textit{ShotAdapter} & \textit{\begin{tabular}[c]{@{}c@{}}Default\\ Dataset\end{tabular}} & \textit{ShotAdapter} \\ \cmidrule{1-5}
\textbf{FVD} & 477.18 & 473.30 & 357.73 & 401.52 \\ \bottomrule
\end{tabular}%
}
\vspace{-7pt}
\end{center}

\end{table}

\section{Motion Filtering}

When curating the training dataset, we aim to select videos with significant motion. To filter such videos, we analyze three types of camera motions: pan (\(t_x\)), tilt (\(t_y\)), and zoom (\(s\)). For each video, we begin by extracting optical flow maps using RAFT~\cite{teed2020raft}, then estimate the homography matrices for each frame using the RANSAC algorithm. Next, we calculate the mean translation vector for the pan and tilt components. For zoom motion, we compute a divergence value that quantifies how much the pixels move towards or away from the frame center. Finally, we average these values across the video to obtain the overall motion magnitude. For \(t_x\), \(t_y\), and scale, we use thresholds of 8, 8, and 0.4, respectively. A video is classified as having significant motion if any of these values exceed the corresponding threshold.

\begin{figure*}[h!]
    \centering
\includegraphics[width=\linewidth]{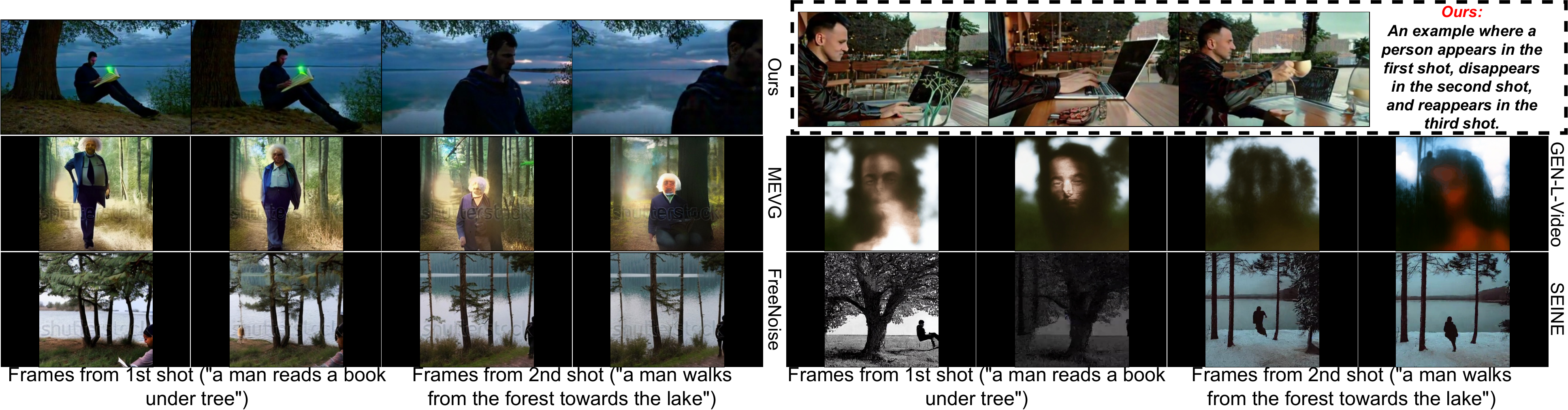}
\caption{\textit{\textbf{Additional qualitative comparison (please zoom-in)}}} 
    \label{fig:qual-comp-re}
\end{figure*}
\vspace{-1.6em}

\section{Implementation Details}
We employ a video diffusion model incorporating joint attention layers within its DiT blocks. The model is fine-tuned using the AdamW~\cite{loshchilov2018decoupled} optimizer with a learning rate of \(5.0 \times 10^{-5}\), weight decay of 0.1, and betas \([0.9, 0.95]\). The learning rate scheduler follows a cosine decay strategy, with 2000 warmup steps, a decay starting at step 2000, with a minimum learning rate of \(2 \times 10^{-5}\). Fine-tuning is performed with a batch size of 128, which is 90\% smaller than the size used during pretraining and runs for 5000 iterations, accounting for less than 1\% of the pretraining iterations, making it a computationally lightweight approach. Note that for all baseline approaches, we use the original pre-trained checkpoint without any fine-tuning. 

\section{Additional Comparisons}

Our objective—motivated by its significance in storytelling and the film industry—is to generate multi-shot videos where shots are separated by (jump) cuts, while ensuring the foreground object remains consistent, regardless of any background changes specified by the user. 
We perform additional comparisons with previous baselines:
(1) \textbf{\textit{SEINE}}~\cite{chen2023seine} focuses on frame interpolation by generating intermediate frames between two source inputs or image-to-video generation, but does not support multi-shot text-to-video generation;
(2) \textbf{\textit{Gen-L-Video}}~\cite{wang2023genlvideo} primarily produces results for video editing that requires a source video, with `multi-text' being used exclusively for editing purposes, restricting the model's ability to generate distinct activities across shots;
(3) \textbf{\textit{FreeNoise}}~\cite{qiu2024freenoise} and (4) \textbf{\textit{MEVG}}~\cite{oh2024mevg} use multi-text to generate a continuous video and are limited in their ability to create `jump cuts' which reduces diversity in camera angles and motion across shots. Additionally, the foreground remains fixed in the same location throughout the video and across scenes. In contrast, our approach enables greater diversity in both aspects.
We compare our approach with all suggested baselines for 2 shots in Table~\ref{tab:comp}. For all but FreeNoise we use the results from the respective webpages since MEVG and Gen-L-Video do not provide code for multi-text prompting, while SEINE only does frame interpolation. For FreeNoise we used our dataset and the official FreeNoise checkpoint.
According to these results (in addition to the qualitative comparison in Fig.~\ref{fig:qual-comp-re}), we outperform all approaches across all metrics by a large margin (this also holds for $>$2 shots for which we do not add results here due to limited space), demonstrating our ability to generate consistent identities under multi-shot video generation settings with rich motion.

We computed the average motion in the datasets and for baselines as shown in Table~\ref{tab:comp}, including continuous generation which uses the original pretrained model. We do indeed see a drop in motion (although we still have more motion than related work), and, after some investigation contribute this to our pre-processed dataset, as the average motion of our filtered dataset is reduced by 38.1\% compared to the original dataset.

\begin{table}[h!]
\centering
\resizebox{\columnwidth}{!}{%
\begin{tabular}{c|cc|cc|cc|cc|c}
            & MEVG & Ours & Gen-L-Video & Ours & SEINE & Ours & FreeNoise & Ours & Cont. Gen \\ \toprule  
Identity Consistency   & 68.5 & \textbf{76.4} & 66.9 & \textbf{82.4} & 69.1 & \textbf{81.9} & 68.2 &  \textbf{86.3} & 88.9 \\
BG Consistency & 69.4 &\textbf{79.8} & 70.9 & \textbf{85.3} & 72.8 & \textbf{87.4} & 77.9 &  \textbf{89.5} & 90.6 \\ 
Avg. Motion & 0.95 & \textbf{1.36} & 1.23 & \textbf{1.25} & 0.75 & \textbf{0.97} & 1.12 &  \textbf{1.19} & 1.42 \\
\bottomrule     
\end{tabular}%
}
\caption{\textbf{\textit{Quantitative comparison with previous works}}}
\label{tab:comp}
\end{table}

\section{ChatGPT Prompt Instruction}

We use ChatGPT to generate our validation dataset which consists of prompts. For each prompt, we provided the instruction: ``Our project involves text-to-multi-shot video generation, where each shot is controlled through local text prompts. I would like you to generate prompts for each video of N shots for 8 videos. For each shot, the background should be XX. Include one human as the foreground object and provide detailed descriptions of the human's appearance." Here, \(N\) corresponds to 2, 3, or 4, and \(XX\) specifies whether the background should be consistent or diverse. After refining the generated results, we finalized the validation prompts used in our experiments. 


\end{document}